# CONTROLLABILITY, MULTIPLEXING, AND TRANSFER LEARNING IN NETWORKS USING EVOLUTIONARY LEARNING

**Rise Ooi**[*]
Hokkaido University
rise.ooi@frontier.hokudai.ac.jp

**C.-H. Huck Yang**[*]
KAUST, Georgia Tech
chao-han.yang@kaust.edu.sa

**Chun-Jen Peng**[*]
National Taiwan University
larrypengcj@gmail.com

**Pin-Yu Chen**
IBM Research
pin-yu.chen@ibm.com

**Vìctor Eguìluz**
IFISC (CSIC-UIB)
victor@ifisc.uib-csic.es

**Narsis A. Kiani**
Karolinska Institute
narsis.kiani@ki.se

**Hector Zenil**
Karolinska Institute
hector.zenil@ki.se

**David Gomez-Cabrero**
Navarrabiomed
david.gomez.cabrero@navarra.es

**Jesper Tegnér**
King Abdullah University of Science and Technology (KAUST), Karolinska Institute
Bioscience, BESE and Computer Science, CEMSE, KSA
jesper.tegner@kaust.edu.sa

## ABSTRACT

Networks are fundamental building blocks for representing data, and computations. Remarkable progress in learning in structurally defined (shallow or deep) networks has recently been achieved. Here we introduce evolutionary exploratory search and learning method of topologically flexible networks under the constraint of producing elementary computational steady-state input-output operations.

Our results include; (1) the identification of networks, over four orders of magnitude, that can compute steady-state input-output functions, such as a band-pass filter, a threshold function, and an inverse band-pass function. Next, (2) the learned networks are technically controllable as only a small number of driver nodes are required to move the system to a new state. Furthermore, we find that the fraction of required driver nodes is constant during evolutionary learning, suggesting a stable system design. (3) Our search technique discovers multiplexing of different computations using the same network. For example, using a binary representation of the inputs, the network can readily compute three different input-output functions. Finally, (4) the proposed evolutionary learning demonstrates transfer learning. If the system learns one function A, then learning B requires, on average less number of steps as compared to learning B from tabula rasa.

We conclude that the constrained evolutionary learning produces large, robust controllable circuits, capable of multiplexing and transfer learning. Our study suggests that network-based and evolution derived computations of steady-state functions, representing either cellular modules of cell-to-cell communication networks or internal molecular circuits communicating within a cell, could be a powerful model for biologically inspired computing. This complements human conceptualizations such as attractor based models, or reservoir computing.

---

[*]The authors conducted this work collaborating with or visiting in Dr. Tegnér's Group.

# 1 Introduction

A Turing machine (TM) is a universal mathematical model of computation. Given a table of rules and a tape, a TM is essentially a discrete system defining a function to be computable if and only if it can be computed on a TM [1]. In practice, a concept of learning is executed as an algorithm which in turn is realized in an architecture. For example, inspired by neural circuits, realized as (deep) neural feed-forward network models of computation [2], learning acts on the connections. Attractor dynamics is another popular conception of neural [3] computations where the fixed points correspond to memory states or to cell-types as in the case of molecular (gene) based computations [4].

Here we are inspired by the computational processing power in cells in living systems, where networks of molecular components collectively transform input signals to the cell into outputs at the cellular level. In the case of neurons, we can represent such a computation as a threshold computation using a ReLU or softmax unit. Yet, we do not understand the underlying network of molecular computations occurring within different cells, realizing different input-output transformations. In this paper, we ask if using an evolutionary search to identify networks capable of computing different input-output functions. Specifically, we address whether such a system can represent different input-output computations in the same network, whether the computations are controllable in an engineering sense, and if transfer learning can occur under such constraints. We formulate a meta-functional networks learning (mFNL) scheme with adopted gradient-based optimization [5], a Simple Genetic Algorithm (SGA) [6] and three milestone Evolutionary Algorithms (EAs): Covariance Matrix Adaptation Evolution Strategy (CMA-ES) [7], Standard Particle Swarm Optimization (SPSO2011) [8], and Ant Colony Optimization for Continuous Domains (ACOR) [9]. Hence, we search for new architectures for computing under the soft constraint that the networks should incorporate a couple of fundamental primitives defined as specified input-output functions. While machine learning techniques such as transductive and inductive learning dealing with graphs have recently become mainstream research focus on artificial intelligence and machine learning [10, 11, 12], we have still limited understanding of the space of putative network architectures for computing. The generative large network models, identified by the evolutionary search, hold the promise to be of broad interest since networks are used as fundamental building blocks for representing data, and computations in biology, social sciences, and communication networks [13]. Our work is predicated on the notion that by imposing fundamental computational operations (input-output transformations), we can generate underlying ensembles of networks realizing such computations.

# 2 Methods - Search for Learning Designer Computing Networks

**Node Dynamics:** To compute input-output functions from a network, we equip the intrinsic network structure with a set of dynamical equations, defining how a given node interacts with its neighbors. The form of such equations is qualitatively motivated by chemical reactions and molecular regulatory control systems such as genes in a cell. A non-linear threshold function summing input from nearby nodes defines the node dynamics as:

$$\frac{dy_i}{dt} = k_{1i}(f(\sum_{ij} W_{ij} y_j) + I_i) - k_{2i} y_i \qquad (1)$$

For each node $i$, $y_i$ represents the activation or expression rate, $f$ is a nonlinear sigmoid function: $f(x) = \sigma(x) = \frac{1}{1+e^{-x}}$, $I_i$ is the input matrix prescribed into the system, and both $k_{1i}$ and $k_{2i}$ are constant vectors which represent the maximum expression and kinetic degradation rate respectively. Note that the $W_{ij}$ represents the influence from nodes $j$ to nodes $i$. For simplicity, we have set all elements of $k_{1i}$ and $k_{2i}$ to 1 for all of our experiments. The motivation originates from the Hopfield network [3], where we follow the modifications as described in previous works [14, 15, 16, 5]. We solve the ODE numerically using the forward Euler's method in which $y_{n+1} = y_n + g(y_n, t_n)\delta t$ and $\delta t$ is set to 1 (we have checked that the results do not change for smaller values of $\delta t$). A network of such connected units collectively computes a given steady-state input-output function. Equation 1 represents a special version of a recurrent neural network [17] and the fully connected network structure looks deceptively similar to a Boltzmann Machine [18] in which the two visible nodes are the input and output nodes respectively while all other hidden nodes (equipped with the above dynamics) are the supporting nodes. However, the links are directed and mutually influencing between nodes for learning appropriate links in order to implement the desired input-output function from the system, see Figure 1a.

**Loss function for computational input-output functions:** We use an "L2-norm [19] error function between the output level points of the simulated function and the desired function". Then, if the simulated function is close to the desired function – if the error is smaller than the tolerance – it is considered to be learned. The loss function $L$ is $L = \sum_{i=1}^{b} \frac{1}{2}(y - \hat{y})^2$ where $y$ is the simulated function's output level, and $\hat{y}$ is the desired function's output level. The maximum tolerance level is set at **5%** of batch size $b$, which is set at 32 data points. We search for any network that computes the desired function, so there is no training set nor test set[2]. Our procedure in practice is as follows: to simulate a network, the "input" is the $W$ weight graph, the "initial state" is a random graph with uniformly distributed weights in [0,1], the "final state" is the learned graph, and the individual "states" refer to final function output values $y$ of each node in the network. To obtain the final outputs $y$ of all nodes, they will have to be first simulated, by solving Equation 1. Before each simulation, the input weights are modified by an optimization algorithm to obtain a better learned simulated graph. If the simulated function outputs $y$ has a low enough tolerance level to desired function outputs $\hat{y}$, then we consider it learned.

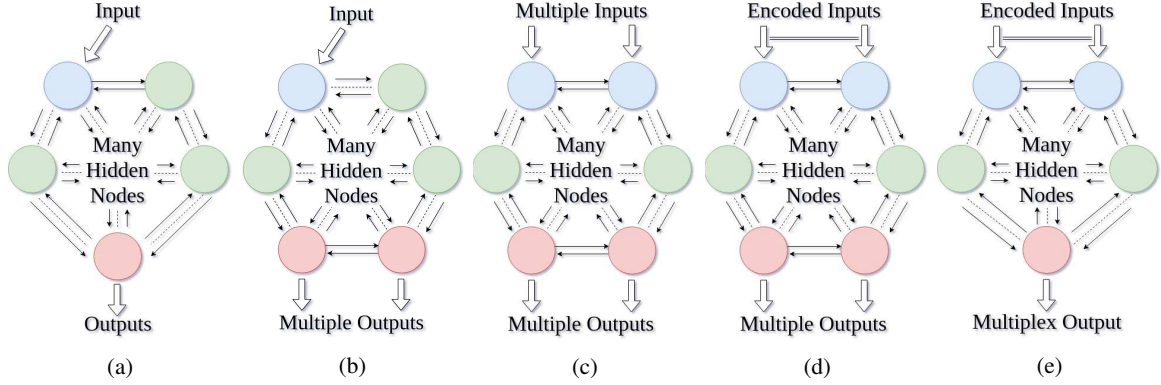

Figure 1: Representation of directed functional network where blue nodes are the visible input nodes, red nodes the visible output nodes, and green nodes the hidden nodes. 1a shows a representation of a standard functional network with an input node and an output node. The dotted lines represent the existence of hidden green nodes not drawn in the figure. 1b shows another representation with a single input node and multiple output nodes. 1c shows another representation with multiple input and output nodes. 1d shows another representation with multiple binary encoded input nodes with multiple output nodes. Finally, 1e shows another representation with multiple binary encoded inputs and a single multiplex output node.

### 2.1 Multiplexing: Learning several input-output in the same network

Next, we generate networks such that any single of them can receive multiple injection inputs and produce multiple output functions. We consider the following four cases: **(1)** A network with 1 node receiving the input and multiple nodes reproducing any desired output from a set of functions. See Figure 1b; **(2)** A network with $N$ nodes receiving the inputs and $N$ nodes reproducing any desired output functions. See Figure 1c; **(3)** A network with $N$ nodes receiving binary encoded inputs and $2^N - 1$ nodes reproduce any desired output functions. See Figure 1d; **(4)** A network with $N$ nodes receiving binary encoded inputs and 1 node reproducing $2^N - 1$ number of any desired output functions. See Figure 1e. For the **(3)** and **(4)** scenarios, we essentially encode the activation of the input nodes using a binary representation, whereas, in scenario **(4)**, the single multiplex output node is intrinsically multi-functional.

### 2.2 Meta-Functional Networks Learning

Searching an optimal network by minimizing the loss function with gradient descent often converges at a local minimum. In contrast to gradient descent, EAs[3] are stochastic, derivative-free methods for non-linear or non-convex continuous optimization. The pseudocode of the Meta-Functional Networks Learning Algorithm (mFNL) and the details of different implementations of utilizing different EAs with mFNL can be found in the appendix. As a reference to the different search algorithms described below, we used **Random Search** as in [21].

**Simple Genetic Algorithm:** Following [22], our SGA uses truncation with elitism method and serves as a baseline for gradient-free algorithms. During initialization of SGA, an initial population of $P$ of random individuals is constructed. A selection pressure $s_p$ is applied to the population to keep only the top $P/s_p$ % of the population with a smaller cost.

---

[2]As an empirical study, we use gene regulatory networks clustering data from [20] to validate our results with finite functional motifs in the appendix

[3]In our experiments, SGA is adopted in the experiments to set a baseline for gradient-free optimization algorithms. Three milestones EAs: CMA-ES, SPSO2011 and ACOR are chosen to enhance optimization capability with competent algorithms under our framework. These EAs are designed to tackle exploration-exploitation trade-off and possess strongly to find potential global optimum on the non-convex cost landscape.

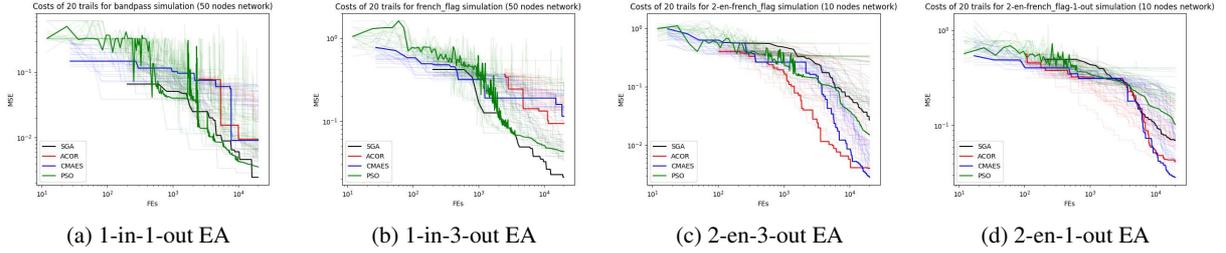

(a) 1-in-1-out EA  (b) 1-in-3-out EA  (c) 2-en-3-out EA  (d) 2-en-1-out EA

Figure 2: 2a, 2b, 2c, and 2d show the mean square error versus the number of function evaluations of 20 trials of optimization by different EAs.

The top $P/s_p$ individuals are then cloned to be the new population $P$. Each individual's gene then mutates with a mutation rate of $p_m$. The algorithm stops when the maximum generation is reached or the population converges, that is, when the error is smaller than the tolerance. Elitism is applied to ensure the best individual remains identical in the new population after mutation, and as explained in [23], the crossover process is omitted since it might disrupt effective structures for our neuroevolution case.

**Covariance Matrix Adaptation Evolution Strategy:** The normal distribution is widely observed in nature. CMA-ES [7] generates isotropic search points, which do not favor any direction, from a multivariate normal distribution. The algorithm updates three parameters, mean, step-size and covariance matrix, to estimate the best multivariate normal distribution that covers the optimal solution near its mean position for a given loss. For CMA-ES, we set the initial population as $4 + \mathbf{floor}(3 * \mathbf{log}(\mathbf{D}))$ for a D-dimension problem[4].

**Standard Particle Swarm Optimization:** Particle Swarm Optimization (PSO), a swarm intelligence algorithm simulating the foraging behavior of bird flocks, searches for the optimal solution in a given search space by updating the velocities and positions of each particle according to the historical experience and the information links defined by the neighborhood topology. The Standard PSO 2011 [8], with a clear implementation that follows universal principals of PSO, is designed and implemented to serve as a milestone for future comparison.

**Ant Colony Optimization for Continuous Domains:** Ant Colony Optimization (ACO), proposed to solve combinatorial optimization problems like scheduling and routing, mimics the pheromone deposition of ants along with the trial, a form of indirect communication. ACOR [9] extends this idea to a continuous domain by estimating the discrete probability distribution with a Gaussian kernel in each dimension. The Gaussian kernel, serving as the pheromone of ants, is updated with the top-k solution archive, to ensure that future samples are constructed near the best solutions and ultimately guides the colony toward the optimum.

### 2.3 Controllability

To formally validate the effectiveness of the generated large-scale networks and multifunctional networks concerning their controllability, that is to discover the input settings where the network can be driven from any initial states to any desired final state within a finite time, we used two widely used schemes such as inspecting the degree distribution of a whole linear – or linearized nonlinear – complex time-invariant network. The first method, by [25], is referred to as a Linear Nodal Dynamics Method (LNDM). Referring to Equation 1, LNDM is simply, when the controllability factor $C_{LNDM} = (I, WI, W^2I, ..., A^{N-1}I)$ has full rank. In other words, a system is "fully controllable" when rank$(C_{LNDM}) = N$. While the second technique, by [26], captures the dynamical process occurring on edges of a directed complex network, is referred to as the Switchboard Dynamics Method (SBD). It substitutes $W = (M - k_2)$ and $I = f(I)$ where our system is "fully controllable" when the controllability factor $C_{SBD} = (f(I), (M - k_2)f(I), (M - k_2)^2 f(I), ..., (M - k_2)^{N-1} f(I))$ has full rank. Here, $M$ is the adjacency matrix of the original digraph's line digraph, $k_2$ is the kinetic degradation rate from Equation 1, and $f(I)$ is the sigmoid-ed version of $I$.

## 3 Learnable large functional networks with targeted feedback

As a baseline reference for our different search algorithms, we used Random Search (RS) to identify solutions such that the networks could compute the desired function. Interestingly, RS could not converge to generate any desired input-output function in all sizes from 3 nodes up to 3000 nodes. This result did not appear to depend on our specific node transfer function as we could replicate the non-convergence using a sigmoid. One possibility is that the search space is merely large to permit a random search to find our target networks, thus suggesting that such computational networks are sparsely distributed in the space of networks.

---

[4]The experiments utilize the official package $pycma$ [24]. See Appendix.

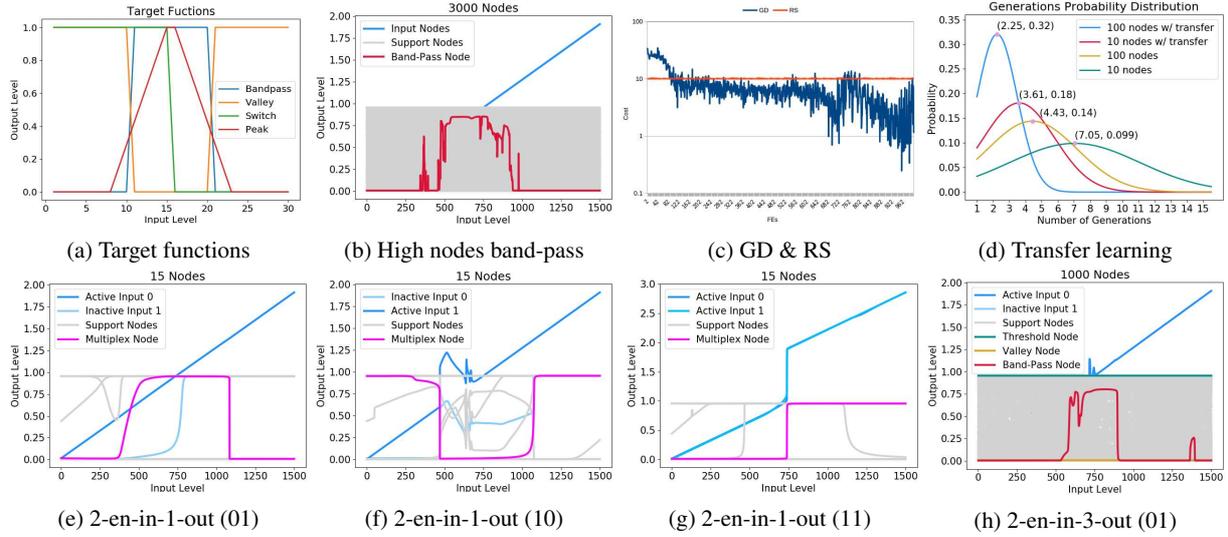

Figure 3: Graphs representing the simulated activity levels within the networks plotted as the output level versus input level. 3a shows the three target functions (band-pass, valley and switch), and also the peak function design used in the transfer learning experiment. 3b shows a 3000 node network outputting the band-pass function. 3c shows the cost against function evaluations of RS and GD. 3d shows the summarized result from the transfer learning analysis experiment. On the second row, 3e, 3f, and 3g show 3 of the 4 simulated activity levels produced by 4 different binary encoding inputs to generate 3 different outputs – band-pass (01), valley (10), and threshold (11) – on a single multiplex output node. The (00) encoding produced no active output and is omitted here. 3h is the 1000 nodes version of the 2 encoded inputs 3 outputs scenario, outputting band-pass (01 encoding).

**GD vs RS:** Inspecting the results shown in Figure 2e-h, each EAs' initial cost starts from 1, and from Figure 3c, we observe that the even the elite of RS has a constant cost of about 10, but GD's cost started from 26 and moved down to 1 much later. The fact that GD started from a much higher cost at the beginning shows that luck is in play here where the GD may not converge when it starts from a "bad" position, which is discussed much greater in detail in the paper [27]. In contrast, with EA we were able to generate stable and controllable networks from 3 nodes up to 3000 nodes, each network targeting a band-pass function[5]. For the case of 3000 nodes, see Figure 3b. It was unexpected, given the failure of random search, that it is actually feasible to find large networks which could robustly compute a defined input-output function despite the large degree of freedom within the network[6]. In our experiments, the maximum node count appears only to be limited by the memory. We are currently exploring what could be achieved using modern evolutionary algorithms [28, 29, 30] and data storing techniques [31, 32, 33]. Such distributed computing can most likely increase the size of the networks to be explored by an additional couple of orders of magnitudes.

### 3.1 Multiplexing of different input-output functions using the same network:

Next, we ask whether we can evolve multi-functional networks which can compute more than one single input-output function. Moreover, we investigated four particular input-output configurations as described in the methods section. For these experiments, the three output functions we tested were the band-pass (French Flag), valley (reversed French Flag), and threshold (Binary Step). We have arbitrarily chosen these functions and we emphasize, any desired function can be approximated with enough nodes. We have particularly chosen the French Flag for illustration because of its significance and difficulty to generate compared with easier functions like Linear.

**Multiple input-outputs scenario:** Figure 1b illustrates the case where a single input can control 3 different outputs while figure 1c shows 3 inputs controlling 3 different outputs. Here the EA learned the three different desired functions respectively. After careful examination comparing 3-in-3-out configuration, we found that the other two input nodes were dummies, and only one input node was lifting all the heavy work. For more details, please refer to Appendix Figure 2.

---

[5] For reproducibility, please refer to the source code, results, and other data at https://bit.ly/2YH9naA
[6] All these networks are generated on a 56-cores desktop computer in less than 4 hours.

Table 1: Controllability analysis experiment result. *GD* (Gradient Descent) and GD-Pruned ($GD_P$) rows show the analysis done on networks generated using gradient descent – with and without pruning [5]. The maximum node count available for *GD* is 18 [34]. $e$ is the fraction of edges – number of edges divided by the maximum number of edges. $\overline{e}$ is $e$ averaged over the evolving process. $N_L$ is the number of driver nodes detected according to the Linear Nodal Dynamics Method while $N_S$ is the number of driver nodes detected according to the Switchboard Dynamics Method. $n_L$ and $n_S$ are the fraction of driver nodes – the number of driver nodes divided by the total number of nodes, of both methods. $\overline{n_L}$ and $\overline{n_S}$ are $n_L$ and $n_S$ averaged over the evolving process respectively.

| Exp. | Nodes | $e$ | $\overline{e}$ | $N_L$ | $n_L$ | $\overline{n_L}$ | $N_S$ | $n_S$ | $\overline{n_S}$ |
| --- | --- | --- | --- | --- | --- | --- | --- | --- | --- |
| GD | 3 | 1 | - | 1 | 0.333 | - | 1 | 0.333 | - |
| GD | 18 | 0.961 | - | 1 | 0.055 | - | 3 | 0.167 | - |
| $GD_P$ | 3 | 0.782 | - | 1 | 0.333 | - | 1 | 0.333 | - |
| $GD_P$ | 18 | 0.052 | - | 13 | 0.722 | - | 5 | 0.278 | - |
| $GA_{1,1}$ | 3 | 1 | 1 | 1 | 0.333 | 0.333 | 1 | 0.333 | 0.333 |
| $GA_{1,1}$ | 3000 | 0.933 | 0.933 | 1 | 0.0003 | 0.0003 | 1441 | 0.483 | 0.486 |
| $GA_{1,3}$ | 1000 | 0.926 | 0.929 | 1 | 0.1 | 0.1 | 474 | 0.474 | 0.476 |
| $GA_{3,3}$ | 1000 | 0.925 | 0.927 | 1 | 0.1 | 0.1 | 473 | 0.473 | 0.472 |
| $GA_{2E,3}$ | 1000 | 0.918 | 0.919 | 1 | 0.001 | 0.001 | 480 | 0.482 | 0.485 |
| $GA_{2E,1}$ | 1000 | 0.931 | 0.931 | 1 | 0.001 | 0.001 | 475 | 0.475 | 0.483 |

**Encoded inputs and multiplexing scenario:** For the following two experiments, we have essentially "encoded" the input nodes in binary format. That is, if both input nodes are not being injected with stimuli, it will be "00" which the least significant number is on the right side. If only the first input node is active, then it should be "01"; if only the second node is active, then it is "10"; and finally if both input nodes are being pumped with injections, then it is "11". We tried these experiments across various nodes count from 5 nodes up to 3000 nodes[7]. Now, figure 3h shows the "01" encoding generating a band-pass function for the first output node. Due to space constraint, not shown here are "10" encoding generating a valley function for the second output node; and "11" encoding generating a threshold function for the third output node. Finally, the last experiment uses the same binary encodings but to output, all the three functions on a single multiplex output node[8]. At "00" encoding, all nodes are essentially inactive and dead. Figure 3e, figure 3f, and figure 3g shows the "01", "10", and "11" encodings successfully generated three functions respectively on the same multiplex output node with the intrinsic multi-functional property.

**The learned networks are controllable:** Here we ask if these large complex networks are amenable to an engineered control or not. To this end, we used the framework of the established controllability analysis as the Linear Nodal Dynamics Method (LNDM) and Switchboard Dynamics Method (SBD) to compute controllability. Table 1 summarizes all the results from the controllability analysis experiment. The number of required driver nodes using LNDM turned out to always be 1, suggesting that the generated networks can be controlled using only one driver node. In contrast, using the SBD, revealed an increased number of required driver nodes as the node count increases. **16.1%** to **48.3%** of the nodes were needed in order to control the network. The average number of required edges and driver nodes as the network evolves, linger close to the final result, for both methods respectively.

## 4 Transfer learning adaptation between functional networks

To quantify transfer learning we used the required number of EA generations while comparing the number of generations to train a steady-state input-output function with the training is initiated from another steady-state function as compared to being initiated from tabula rasa. The intuition being that the intermediary state may already have some structural properties - or network motifs - facilitating the learning of another function. For the reference case, we compute the required generations to achieve the band-pass function from tabula rasa. For transfer learning case, we first compute the required evaluations from tabula rasa to the intermediate state "the peak function" (Figure 3a), then from peak state to the final band-pass form.

---

[7] The number of nodes is chosen arbitrarily for the four configurations described above. For large-scale analysis purposes, we have also tested out with 1000 nodes for all four configurations included in Table 1.

[8] The term "multiplex" refers to the capacity of a network to compute more than a single steady-state input-output function from the same input source.

Controllability, Multiplexing, and Transfer Learning in Networks using Evolutionary LearningAs EA generations can differ significantly between different epochs, we average the estimations using at least 100 networks for each configuration. Figure 3d shows the probability distribution of the number of generations required for EA to learn to output the band-pass function. For simplicity, the transfer learning results from two separate processes – tabula rasa to peak, then peak to band-pass – are merged into one. With a batch size of 32 and a population size of 224, the average number of generations $G$ required from the transfer learning experiment is 3.61 (N=10) and 2.25 (N=100). In contrast, training from tabula rasa straight to band-pass has larger $G$ of 7.05 (N=10) and 4.43 (N=100), a significant difference. Also, the elite of a population of randomly generated networks has an average cost of 10.184 (N=10) and 10.134 (N=100); whereas the networks that are initiated from the peak has an average cost of 7.824 (N=10) and 6.876 (N=100). This effect is validated across a various number of nodes. The generations needed to reach band-pass from the peak is distinctively lower than to train a band-pass from tabula rasa, thus supporting transfer learning as defined here.

**Stability analysis:** To further assess the existence of transfer learning we evaluated and compared the stability between a network representing one function to the case of another function represented by different network. To this end we classified the equilibria of a 2-D dynamical system [35, 34], by computing the trace $\tau = tr(J) = \frac{\partial f_1}{\partial x_1} + \frac{\partial f_2}{\partial x_2}$ and determinant ($\delta$) of the Jacobian matrix $\Delta = det(J) = \frac{\partial f_1}{\partial x_1}\frac{\partial f_2}{\partial x_2} - \frac{\partial f_1}{\partial x_2}\frac{\partial f_2}{\partial x_1}$ where $f$ is the sigmoid function. Compared to networks generated by the gradient [5] or evolutionary-gradient [34] based methods, the determinants of $J$ using EAs are **37.63%** larger, which means stabler systems for nodes count 3 to 18. At 3000 nodes, $det(J)$ becomes **29.58%** larger and $tr(J)$ is **12.77%** smaller than 1000 nodes', which means the large systems generated by EAs to be used in transfer learning experiments are increasingly stable[9].

## 5 Discussion

The work presented here can be viewed as bridging between learning fundamental computational input-output computation without fixating the topology of the graph in advance and is clearly motivated by a perspective of how computing could occur in living systems. For instance, functional connectivity or strength of synchronous activity between typologies reveals biological changes (e.g., disease-related) in brain physiology [36, 37] and gene regulatory networks studies. Previous approaches to swarm robotics also require a high demand to design functional sensors networks when large systems possess sensors that have been used for the detection of multiple signals [38]. Here, we encapsulate computations using a fine structure, i.e. network structure and dynamics, which in turns performs well defined input-output transformations of external signals hiding the internal complexity from the observer via the encapsulation. Using evolutionary algorithms, we find here that such systems can readily evolve, and learn both structure and function. Our embedded computational network architecture can be interpreted either at the level of cells (an internal network of molecules such as genes, proteins, metabolites) or alternatively as group or module of cells effective performing input-output transformation of external signals relative to the module. Naturally, this does not exclude a nested architecture ranging from cells encapsulating molecular networks to interacting modules – composed of cells.

Historically, it is a recurrent theme to investigate internally stable states in networks of interacting elements equipped with some dynamics. Examples include pure spin glass models [39] or versions thereof such as attractor networks (e.g. Hopfield [3]), networks of McCulloch-Pitts neurons [40], or Boolean networks representing either molecular circuits [4] or interacting cells. This important line of work essentially frames the problem such that; given a network architecture (e.g., symmetric connectivity matrix, statistical connectivity constraints) and rules for local computation (e.g., Boolean rules, threshold dynamics, a summation of inputs) then what are the stable states, transitions, and dynamics within the system. Recent progress has demonstrated that deep neural networks can be trained as efficient statistical classifiers, and it has been known for decades that even shallow neural networks are universal in a Turing sense.

Now, in contrast, our work flips the problem formulation around in the sense that we ask how could a system of interacting components realize a given computation defined as an input-output transformation. This is similar in spirit in addressing how to design a transistor or a chip to implement in some medium an inevitable signal transformation. What then are the advantages and limitations of our approach? Using a global constraint, such as a computation of an input-output transformation, we open up the possibility to assess what design principles, if any, are necessary or sufficient to realize such computations. Such computational design principles can readily be searched for in the increasing amount of molecular data defining networks of genes, proteins, and metabolites. From a biological standpoint, there has been a surge of studies asking how adaptation, detection of fold changes and French-flag (i.e., band-pass) can be realized in biological systems [41, 42]. However, as rule, this has been investigated using strong constraints on the specific form of dynamics and exhaustive search in small 3-node circuits. Here we introduce the notion to use evolutionary search algorithms in order to access large (thousands of nodes) networks. Our findings demonstrate the existence and fast convergence to these large systems instantiating the steady-state computations we investigate in this paper. Interestingly, in contrast to gradient descent techniques where we find a limit in terms of one-order of magnitude networks ($N < 20$) [5], an evolutionary search can exploit a much large architectural space. Thus, we can, therefore, address a computational analog to the large networks we observe inside cells or between a large number of cells. Finally, we like to remark that in our gradient-based experiments, gradient descent receives only information before and after simulation, and does not interfere the ordinary differential equation simulation/solving process (see Equation 1). Under this circumstance, the computed gradients are fuzzy, and this is why gradient-less optimization algorithms may perform better here. Interestingly, [43] describes solving ordinary differential equations using scalable backpropagation, which we intend to explore further in searching for novel computational architectures.

---

[9]See Appendix section B. for dynamics analysis.

We like to remark that our approach may appear akin to reservoir computing [44] but there are important differences. First, here we do not employ gradient descent training of the local (reservoir) network, since the evolutionary search benefits from explorative search beyond local minima. Secondly, we train the system to learn steady-state functions, whereas liquid computing or reservoir computing exploits transients, i.e. the temporal dynamics in the system, without necessarily requiring stable states. Yet, the analogy between steady-state and temporal computing will be further investigated in future studies where we will extend our analysis to transient temporal signals. Furthermore, in our work, we demonstrate for the first time, to the best of our knowledge, multiplexing in such large network architectures. Hence, a network can be evolved to learn several different input-output transformations. Moreover, these can be configured in different manners, such as allowing, for example, a binary input coding projecting to either one or several different output channels. One possible disadvantage with such large systems, compared to smaller computational network modules would be that the issue of control in an engineering and computational sense could become insurmountable. However, when computing controllability for our different circuits, we find that the circuits are indeed controllable in a technical sense, which could be a consequence of their fairly dense wiring.

Finally, one advantageous feature could be that when the system is trained for one task, the system is thereby conditioned to learn other tasks faster. This was in part previously observed for the reinforcement learning networks that were trained on Atari games. Hence, for any learning system, it would be desirable that transfer learning comes out as a natural feature. We quantified this phenomenon by demonstrating that convergence was on average much faster provided that the networked learned one nearby input-output transformation.

...Controllability, Multiplexing, and Transfer Learning in Networks using Evolutionary Learning

# Appendix

## A.1. Algorithm detail

For a dynamic network with N nodes, our goal is to optimize the adjacency weight matrix with $N^2$ variables. We define the loss function as the mean squared error between the target function and the simulated output. We calculate the simulation, following the forward Euler's method described in section 2.1, with the given weight matrix and one or multiple linear inputs, depending on the experiment network structure depicted in Fig 1. We use different EAs to find the weight matrix which best fit the target function with minimum MSE loss.

---

**Algorithm 1** Meta-Functional Networks Learning Algorithm

**Input:** max generations $G$, points batch size $B$, population size $N$, truncation number $T$, fitness function $F$, tolerance error $c$, mutation rate $\delta$, convergence criteria $C$, selection function $\Upsilon$, mutation function $\Psi$, random initialization function $\Re$.
$\Re(P)$;     // Randomly initialize N networks
**for** $g = 1, 2, ..., G$ generations **do**
  **for** $n = 1, 2, ..., N$ individuals **do**
    $c_n = F(P_n, B)$;     // Simulate ODE and compute f as Section 2.1
  **end for**
  **if** $c_{min} < C$ **then**
    return $P_0$ (Elite)     // Convergence successful
  **else**
    $P = \Upsilon(P, f, T, B)$     // **Selection** for different EAs
    $P = \Psi(P, \delta)$     // **Mutation** for different EAs
  **end if**
**end for**

---

## A.2. SPSO 2011

In SPSO 2011, the swarm size is suggested as 40 and each particle $i$ is initialized with a random position $x_i$, velocity $v_i$, previous best position $p_i$, and best position of all the previous best positions in the neighborhood $l_i$: where

---

**Algorithm 2** SPSO 2011 on Functional Network

$x_i(0) = U(min_d, max_d)$,
$v_i(0) = U(min_d - x_{i,d}(0), max_d - x_{i,d}(0))$,
$p_i(0) = x_i(0)$,
$l_i(0) = argmin_{j \in N_i(0)}(f(p_j(0)))$.

---

**U(min_d, max_d)** is a random number drawn from a uniform distribution within $[min_d, max_d]$, and $N_i(t)$ is a set of neighbors of particle $i$ at time $t$ defined by the swarm topology.

The SPSO 2011 velocity update equations eliminate the coordinate dependency by creating a hypersphere according to $x_i$, $p_i$, and $l_i$. The hypersphere is defined as:

$$H_i(G_i, ||G_i - x_i||),$$

with center $G_i$ and radius $||G_i - x_i||$. When the personal best $p_i(t)$ is not the neighborhood previous best $l_i(t)$, the center $G_i$ is defined as:

$$G_i = \frac{1}{3}(x_i + (x_i + c(p_i - x_i)) + (x_i + c(l_i - x_i))) = x_i + \frac{c}{3}(p_i + l_i - 2x_i).$$

If the personal best is the best in the neighborhood, *i.e.* $p_i(t) = l_i(t)$, the center $G_i$ is defined as:

$$G_i = \frac{1}{2}(x_i + (x_i + c(p_i - x_i))) = x_i + \frac{c}{2}(p_i - x_i).$$

In both cases, the acceleration constant $c$ is:

$$c = \frac{1}{2} + \ln(2) \simeq 1.193.$$

A random sample $x_i^{'}$ is drawn from the hypersphere with uniform random direction and uniform radius:

$$r = U(0, ||G_i - x_i||).$$

Therefore, the velocity update equation and the new position of SPSO 2011 are defined as:

$$v_i(t+1) = wv_i(t) + x_i^{'}(t) - x_i(t), x_i(t+1) = x_i(t) + v_i(t+1) = wv_i(t) + x_i^{'}(t),$$

where the inertia weight is:

$$w = \frac{1}{2\ln(2)} \simeq 0.721.$$

**A.3. Ant Colony Optimization**

For ACOR, the initial solution construction process proceeds as following. After evaluating $k$ randomly generates solutions, they are stored along with their fitnesses and weights in the solution archive. The solution archive is sorted horizontally according to the fitness so that the best solution $s_1$ is on the top.

For the $i$-th dimension, a Gaussian kernel PDF $G^i(x)$ is constructed with multiple weighted normal distributions $g_j^i(x)$:

$$G^i(x) = \sum_{j=1}^{k} \omega_\ell g_\ell^i(x) = \sum_{\ell=1}^{k} \omega_\ell \frac{1}{\sigma_\ell^i \sqrt{2\pi}} e^{-\frac{(x-\mu^i)^2}{2\sigma_\ell^{i^2}}}.$$

Such a PDF is easy to sample and provides better flexibility to describe the landscape. Three elements decide the shape of the Gaussian kernel:

- $\omega_\ell$ is the **weight** of the solution $s_\ell$.
- $\mu^i$ is $i$-th value of all the solutions in the archive.
- $\sigma^i$ is the standard deviation that determines that new samples.

Following are some parameters used by ACOR that also affects the Gaussian kernel construction:

- $k$ is the size of the solution archive, which works like the memory of the swarm. It should not be smaller than the number of dimensions;
- $q$ represents the locality of the search process, i.e., how often should we not pick the best solution in the archive. It controls the exploration. A larger value of $q$ leads to slower yet more robust convergence;
- $\xi > 0$ represents the speed of convergence, which has an effect similar to the *pheromone evaporation* rate in ACO. It works like the selection and enhances the good solutions in the archive.

For general solution construction, we first define $m$ as the number of ants used in an iteration for sampling. During each iteration, $m$ new solutions are constructed dimension by dimension by sampling the $i$-th dimension Gaussian kernel $G^i$

Pheromone update is accomplished by adding the new samples, their fitness, and the corresponding weight into the archive. Then, the solution archive is sorted according to the fitness and we only keep the top-k solutions in the archive. This selection process ensures only the best solutions are kept in the archive. Therefore, future samples will be constructed near these best solutions and guide the ants toward the optimum.

**A.4. CMA-ES**

For CMA-ES, we set the initial population as $\mathbf{4 + floor(3*log(D))}$ for a D-dimension problem as our submission draft. Comparing to other Evolutionary Algorithms, CMA-ES adopts a relatively small population size, allowing it to update the underlying model frequently and reduce function evaluations

## B. Dynamics analysis for learnable functional networks

A dynamics adjacency matrix $W$ from Equation 1 is said to be asymptotically stable, for any given positive semi-definite symmetric matrix $Q$, there exists a positive-definite symmetric matrix P that satisfies the following Lyapunov criterion:

$$P - WPW^{\mathrm{T}} = Q \qquad (2)$$

Controllability, Multiplexing, and Transfer Learning in Networks using Evolutionary Learningwhere $M$ could be computed by:

$$W = \frac{\partial \vec{F}}{\partial \vec{x}} = \begin{pmatrix} \frac{\partial f_1}{\partial x_1} & \frac{\partial f_1}{\partial x_2} \\ \frac{\partial f_2}{\partial x_1} & \frac{\partial f_2}{\partial x_2} \end{pmatrix} \quad (3)$$

And the convergent solution of x could be computed by:

$$\delta\vec{x}(t) \propto e^{At}\delta\vec{x}(0) \quad (4)$$

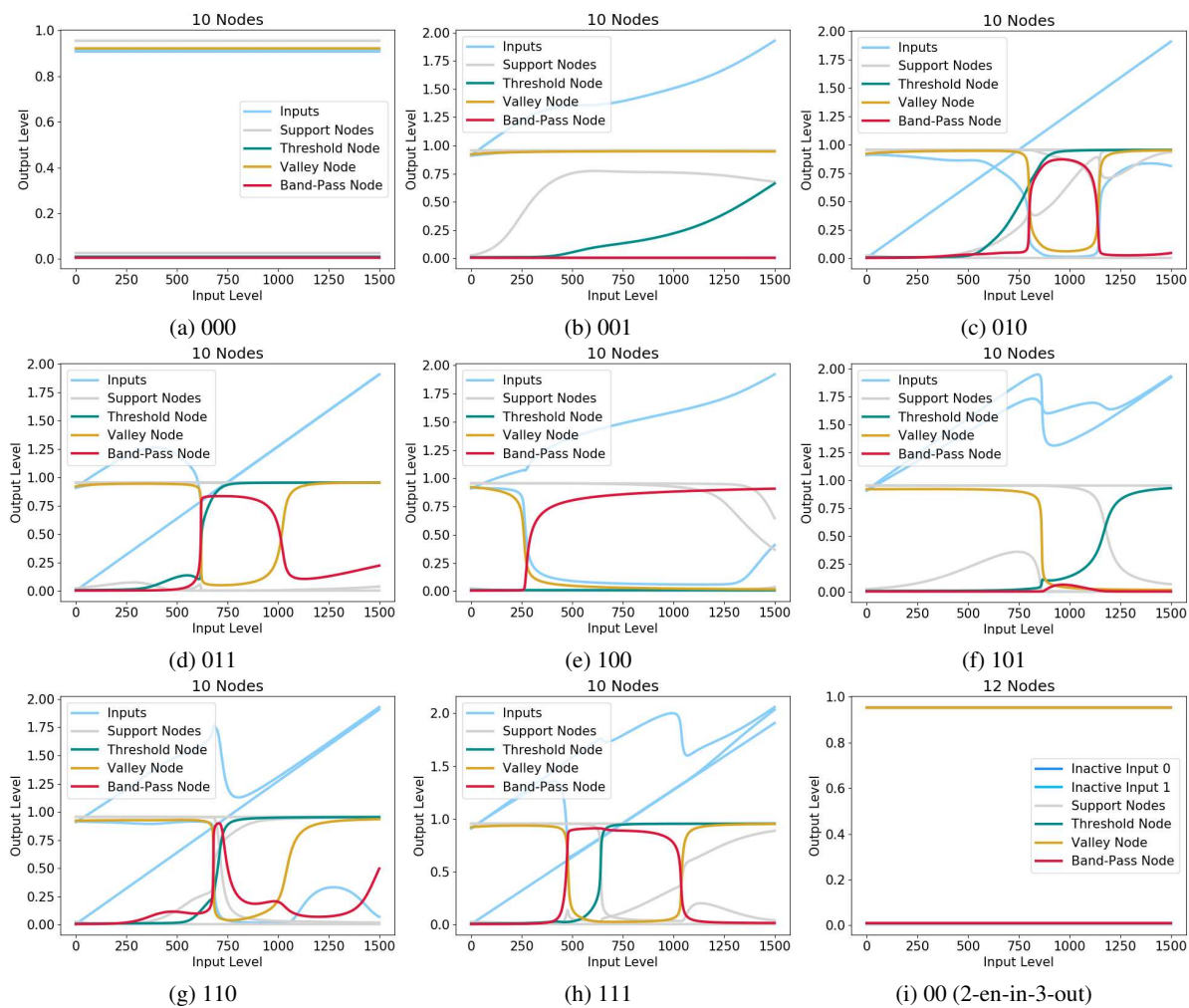

Figure 4: One of the experiment was to train the network to receive 3 inputs and output 3 functions – band-pass, reversed band-pass, threshold – on 3 different output nodes separately. 4h was the final graph, which successfully output 3 correct functions. However, since receiving 3 inputs is just the "111" encoding, we wondered what other encodings (000, 001, ..., 110) might produce on a network that is only trained for "111". 4a to 4g are the other results. 4i was an addition to show that without input, no matter what their desired functions are, the nodes will not have an active expression. Deeper inspection of the 3 input nodes 3 output nodes network structure revealed that only 1 input node was crucial in determining the output functions (See 4c and 4d) while other nodes have much less influence, agreeing the controllability analysis result (Table 1) from the main text.